\documentclass[sigconf]{acmart}
\pdfoutput=1

\usepackage{tikz}
\usepackage{algorithm2e}
\usepackage{algorithmic}
\usepackage{amsfonts}
\usepackage{colortbl}
\usepackage{hyperref}
\usepackage{listings}
\usepackage{multirow}
\usepackage{pifont}
\usepackage{subfig}
\usepackage{textcomp}
\usepackage{xcolor}
\usepackage{xspace}
\usepackage[font=small,labelfont=bf]{caption}

\hypersetup{
           breaklinks=true,  
           colorlinks=true,  
           pdfusetitle=true, 
        }

\newcommand{\pmoveFull}{Parameter Movement\xspace}
\newcommand{\amoveFull}{Activation Movement\xspace}

\newcommand{\pmove}{PMove\xspace}
\newcommand{\amove}{AMove\xspace}

\newcommand{\name}{MoNDE\xspace}

\newcommand{\keyword}[1]{
    \vspace{0.02in}
    \noindent \textbf{#1}%
}
\AtBeginDocument{%
  }

\setcopyright{none}
\renewcommand\footnotetextcopyrightpermission[1]{}

\begin{document}
\title[MoNDE: Mixture of Near-Data Experts for Large-Scale Sparse Models]%
{
    \vspace{-0.10in}
    \huge
    MoNDE: Mixture of Near-Data Experts for Large-Scale Sparse Models
}

\settopmatter{authorsperrow=1,printfolios=false}
\author{
    \Large
    Taehyun Kim$^{\dag\ddag}$\quad Kwanseok Choi$^{\dag}$\quad Youngmock Cho$^{\dag\ddag}$\quad Jaehoon Cho$^{\dag}$\quad Hyuk-Jae Lee$^{\dag\ddag}$\quad Jaewoong Sim$^{\dag}$
}
\affiliation{
    \vspace{0.03in}
    \institution{$\dag$Seoul National University\quad\qquad$\ddag$Inter-University Semiconductor Research Center}
    \vspace{0.02in}
    \normalsize
    \country{\{taehyunzzz, kwanseok.choi, fudsla, jaehoon.cho, hyukjae, jaewoong\}}@snu.ac.kr
}

\thanks{This is a preprint version. The authoritative version will appear in the Proceedings of the 61\textsuperscript{st} ACM/IEEE Design Automation Conference (DAC), June 2024.} 

\begin{abstract}
Mixture-of-Experts (MoE) large language models (LLM) have memory requirements
that often exceed the GPU memory capacity, requiring costly parameter movement
from secondary memories to the GPU for expert computation.
In this work, we present Mixture of Near-Data Experts (MoNDE), a near-data
computing solution that efficiently enables MoE LLM inference. MoNDE reduces the
volume of MoE parameter movement by transferring only the \emph{hot} experts to
the GPU, while computing the remaining \emph{cold} experts inside the host
memory device. By replacing the transfers of massive expert parameters with the
ones of small activations, MoNDE enables far more communication-efficient MoE
inference, thereby resulting in substantial speedups over the existing parameter
offloading frameworks for both encoder and decoder operations.
\end{abstract}
\maketitle
\pagestyle{plain}

\section{Introduction} \label{sec:intro}

Transformer-based large language models (LLMs) have demonstrated impressive
performance in a variety of natural language processing tasks such as question
answering, machine translation, and even software code 
generation~\cite{bert,gpt4,codeparrot}. This outstanding model performance
can largely be attributed to \emph{unprecedented} model sizes that are
constantly growing over the past years. However, scaling model capacity
inevitably leads to an increase in computational costs and memory requirements,
thereby making it increasingly difficult to train and serve due to limited
hardware budgets, even for many leading companies in the industry~\cite{deepspeed}.

Mixture of Experts (MoE) has gained attention as a method to scale model sizes
\emph{without} proportionally increasing the computation
cost~\cite{outrageously-large,switchtransformers}. In MoE Transformers,
the feed-forward network (FFN) layer in the Transformer is replaced by the
MoE FFN layer that contains multiple \emph{expert} FFNs with a
\emph{gating} network. Because \emph{only} a few experts are selected by the
sparse gating function to perform computation for a given input token, the
computational cost is relatively cheaper than non-MoE models with the same
number of parameters. As such, MoE is beginning to be adopted in production
LLMs, as recently demonstrated in OpenAI's GPT-4~\cite{gpt4}.

Although MoE Transformers can significantly increase model capacity without
proportionally increasing training costs, serving such MoE Transformers for
inference remains challenging because \emph{all expert parameters} still need
to reside in the GPUs, which can be costly for inference serving scenarios.
Existing deep learning frameworks, such as Microsoft's DeepSpeed~\cite{deepspeed},
alleviate memory capacity requirements by offloading model parameters to the CPU
memory or SSDs and bringing them back to the GPU when needed for
computation~\cite{zerooffload,zeroinfinity}. However, such parameter offloading
techniques lead to considerable data movement overhead, which adversely impacts
inference latency. Moreover, the latency of expert transfers cannot be effectively
hidden by computation through well-known techniques such as parameter prefetching.
This is because the expert parameters to transfer are \emph{dynamically} determined
just before the expert FFN computation unlike non-MoE models, based on the input
activations and sparse gating functions. 

\begin{figure}[t] 
 \centering
 \includegraphics[trim=4.2in 3.3in 4.2in 3.2in, clip=True,
                 width=\columnwidth]{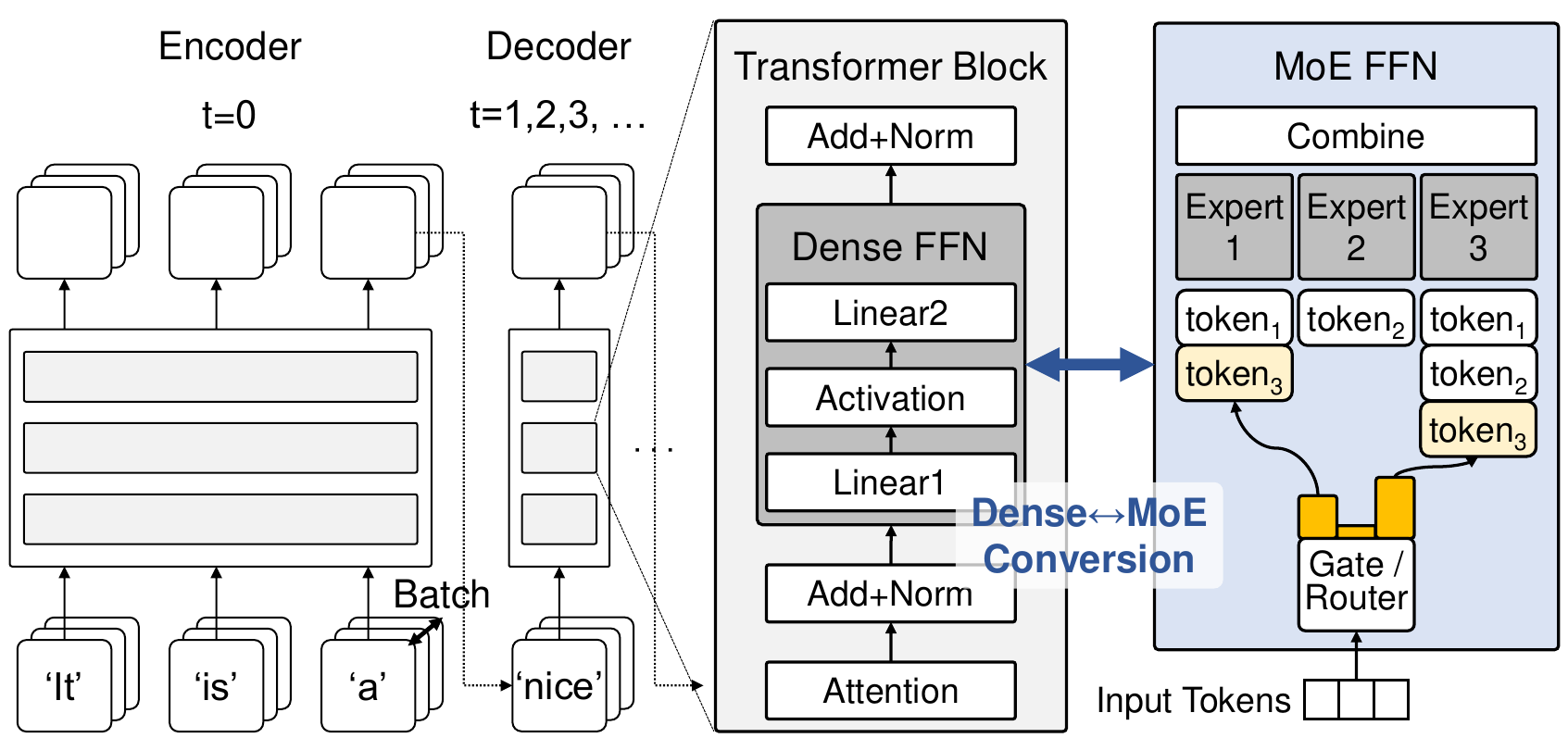}
 \vspace{-0.20in}
  \caption{
      Overview of a Transformer block (left) and an MoE FFN layer (right) with
      $E=3$ experts and top-2 routing.
     \label{fig:moe_background}
  }
 \vspace{-0.10in}
\end{figure}

\begin{figure*}[ht]
    \centering
    \vspace{-0.10in}
    \includegraphics[trim=0.35in 4.3in 0.35in 4.25in, clip=True,
                        width=0.98\textwidth]
                        {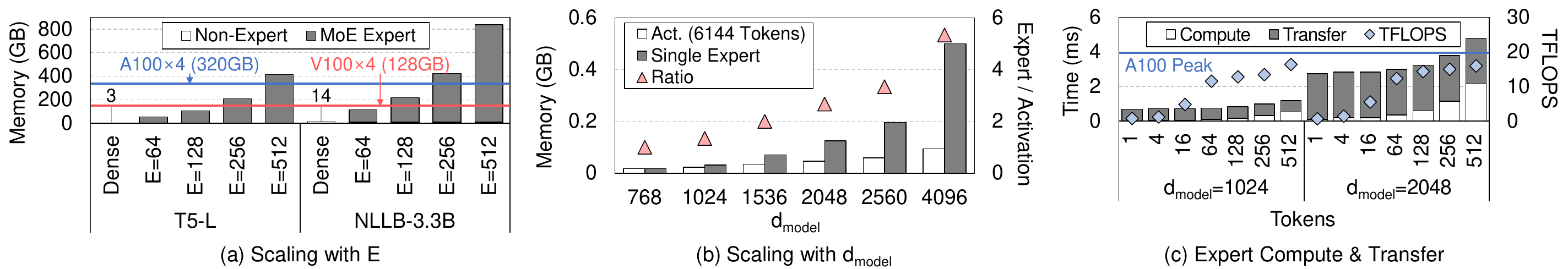}
    \vspace{-0.10in}
    \caption{
        Characterization of MoE Transformers:
        (a) MoE scaling with $E$
        (b) MoE scaling with $d_{model}$
        (c) Latency comparison of computation and transfer of a single expert
        across input token sizes and $d_{model}$ on NVIDIA A100 + PCIe Gen4 $\times$16.
    }
    \vspace{-0.05in}
    \label{fig:moti_analyses}
\end{figure*}

In this work, we present \emph{Mixture of Near-Data Experts (\name)}, a
near-data processing (NDP) solution that efficiently serves MoE Transformers
for inference in a cost-effective way. The key observation of our work is
that the majority of experts in the MoE layer receive a significantly small
number of tokens. Consequently, for these experts, the transfer of expert
parameters to the GPU takes substantially longer than the subsequent expert
FFN computation in the GPU. Furthermore, with only a few tokens to process
for these \emph{cold} experts, the high compute throughput offered by the
commodity GPU is severely underutilized.

Based on the observation, \name{} enables the paradigm of \emph{\amoveFull}, in
which the costly transfers of expert parameters are replaced with
\emph{relatively cheap} activation transfers between the GPU and the host
memory device that contains the \name{} NDP units. With the activations from
the GPU (i.e., outputs from attention layers), the \name{} NDP units perform
expert computations. The resulting output activations from \name{} are then
transferred back to the GPU for subsequent Transformer operations (i.e.,
operations in the attention layers). We also present a novel GPU-\name{}
load-balancing scheme that exploits the \emph{skewed} nature of MoE and runs
expert computations concurrently in both the GPU and \name{} to further reduce
inference latency. Our evaluation shows that \name{} outperforms the existing
expert parameter offloading framework by up to 7.5{$\times$} and 3.7{$\times$}
for encoder and decoder operations with an area overhead of 3.0$mm^2$ for our
\name NDP units. In summary, this paper makes the following contributions:
\begin{itemize}
    \item We propose \name, a near-data processing system that targets MoE
    Transformer inference. To our knowledge, this is the \emph{first} work that
    enables efficient MoE inference by exploiting NDP units.
    \item We provide characterizations of MoE operations and analyze the
    factors that contribute to performance degradation.
    \item We demonstrate the benefit of latency reduction by replacing \emph{\pmoveFull}
    with \emph{\amoveFull} and show the effectiveness of the \emph{GPU-\name{}
    load-balancing} scheme.
\end{itemize}
\vspace{-0.05in}
\section{Background and Motivation} \label{sec:back}

\subsection{Mixture of Experts} \label{ssec:back-moe}
The Mixture of Experts (MoE) is an ensemble technique that aims to enhance
model performance. The key motivation behind using MoE is to increase the model
capacity \emph{without} a proportional increase in computation.
Recent studies show that Transformer-based models that adopt MoE (dubbed MoE
Transformers) achieve substantial improvements in model performance over
conventional \emph{dense} Transformers~\cite{switchtransformers}.

Figure~\ref{fig:moe_background} shows a high-level overview of MoE
Transformers, where MoE is applied to the feed-forward network (FFN) within the
Transformer block. The MoE FFN layer combines multiple copies of dense FFNs,
called \emph{experts}. Each expert comprises two linear layers and an
activation layer in between, which is the same as the FFN layer in conventional
Transformer blocks. The key component in MoE is the gating/routing network,
which determines the experts to which an input token is routed. For each
input token, the gating function computes the probability distribution over
the experts and creates a (token {$\times$} expert) score map that is used to
route each token to the top-$k$ experts. Once the experts process the routed
tokens, the outputs are combined and re-organized into the original order of
the input tokens. These are then forwarded to the subsequent Transformer block.
Table~\ref{table:notations} summarizes the notations regarding the model and
embedding dimensions used throughout this work.

{
    \small
    \begin{table}[h]
    \vspace{-0.05in}
    \caption{Notations for model and embedding parameters.} 
    \vspace{-0.10in}
    \begin{center}
        \small
        \begin{tabular}{|c||c|c||c|} 
            \hline
            \textbf{Term} & \textbf{Description} & \textbf{Term} & \textbf{Description} \\
            \hline
            $B$ & Sequences per batch & $S$ & Tokens per sequence \\
            $E$ & Experts per MoE & $d_{model}$ & Embedding dim. \\
            $d_{ff}$ & Intra-FFN dim. && \\
            \hline
        \end{tabular}
    \label{table:notations}
    \end{center}
    \vspace{-0.10in}
    \end{table}
}

\vspace{-0.03in}
\subsection{Characterization of MoE Transformers} \label{ssec:c-moe}
\noindent\textbf{MoE Parameter Scaling.}
Figures~\ref{fig:moti_analyses}(a) and ~\ref{fig:moti_analyses}(b) illustrate the
parameter scaling trend of MoE models across the number of experts and embedding
dimensions. The parameter sizes of MoE LLMs exhibit an asymptotically-linear growth
in relation to the number of experts and can easily exceed the GPU memory capacity
even for multi-GPU computing nodes. For instance, T5-Large~\cite{t5} requires 
approximately 3 GB of memory, whereas Switch
Transformers-Large~\cite{switchtransformers}, which is \emph{a 128-expert MoE
version} of T5-Large, demands approximately 100 GB (34$\times$).
Scaling $d_{model}$, which is one of the common methods of scaling LLMs,
quadratically increases the gap between the expert parameters and
linearly-scaling activation data.

\keyword{Parameter Transfer Bottleneck.}
As MoE Transformers scale to trillions of parameters, relying solely on GPU
memory for hosting entire model parameters is unlikely to be a viable solution.
Meanwhile, emerging interconnect technologies, such as Compute Express Link
(CXL), enable the addition of tens of terabytes of memory capacity to the CPU.
As such, considerable efforts have recently been made to leverage the large
host memory by offloading model parameters to the CPU memory and fetching them
on-the-fly to the GPU when required for computation~\cite{deepspeed,se-moe}. 
For example, in an MoE-specific offloading approach~\cite{se-moe}, dense
non-expert parameters are permanently stored in the GPU memory, while the
sparse (but massive) MoE expert parameters are offloaded to the CPU due to the
constraints of GPU memory capacity.

However, we observe that \emph{na\"ively} transferring the offloaded expert
parameters back to the GPU becomes a major performance bottleneck for MoE
inference. Figure~\ref{fig:moti_analyses}(c) shows the execution time for an
expert across the numbers of routed tokens and $d_{model}$ sizes, which we
decompose into two components: expert computations and parameter transfers.
The results show that transferring a single expert parameter to the GPU takes
significantly longer than the expert computation, particularly when the expert
receives a small number of tokens (e.g., up to 30{$\times$} longer for a single
routed token).

\begin{figure}[b]
    \centering
    \vspace{-0.10in}
    \includegraphics[trim=0.0in 0.15in 0.0in 0.15in, clip=True,
                         width=0.92\columnwidth]{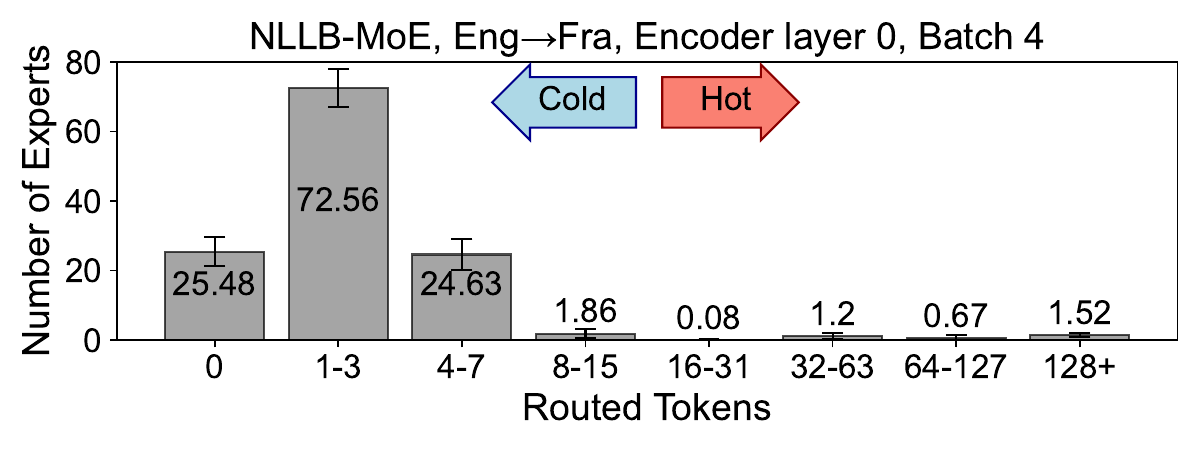}
    \vspace{-0.15in}
    \caption{
        Average token distribution across all inputs for translation task using the
        NLLB-MoE model and FLORES-200 dataset.
    }
    \vspace{-0.10in}
    \label{fig:moe_dist}
\end{figure}

\begin{figure*}[t]
    \centering
    \vspace{-0.10in}
    \includegraphics[trim=1.8in 3.5in 2.0in 3.5in, clip=True,
        width=0.96\textwidth]{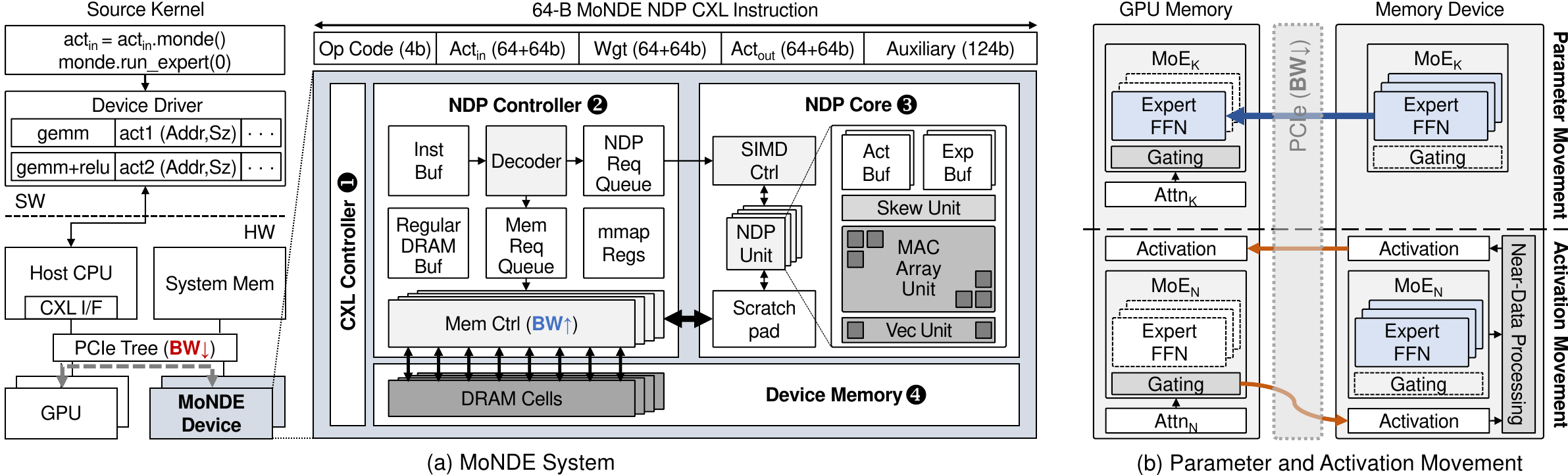}
    \vspace{-0.10in} 
    \caption{
        \small
        \name{} overview:
        (a) \name{} system
        (b) \pmoveFull and \amoveFull.
    }
    \vspace{-0.10in} 
    \label{fig:monde_overview}
\end{figure*}

\keyword{Expert Skew and Load Imbalance.}
As discussed in Section~\ref{ssec:back-moe}, the gating network determines the
expert to which a token is routed. We observe that the number of tokens that
each expert receives substantially differs across the experts in the MoE layer.
Figure~\ref{fig:moe_dist} illustrates the imbalance in the distribution of
tokens routed to experts, with the x-axis representing the number of routed
tokens and the y-axis indicating the number of experts for each token range.
Only a small number of \emph{hot} experts process a large number of tokens,
while the majority of the remaining \emph{cold} experts process significantly
fewer tokens (e.g., 0-7 tokens).
This implies that the operational intensity (i.e., compute-to-memory ratio)
varies among experts, with the majority of expert computations being
memory-bound, thereby leading to severe underutilization of GPU cores.

\subsection{Opportunities for Near-Data MoE} \label{sec:back-opportunity}
The discussion in Section~\ref{ssec:c-moe} implies that specialized hardware
with lower peak compute could provide performance similar to the GPU for expert
computation. In addition, the overall execution time of MoE can even improve
over using the GPU with expert transfers if the parameter movement could be
avoided, which motivates our work. To this end, we design specialized NDP
hardware for effectively processing \emph{fat and wide} matrix multiplications
of cold experts (Section~\ref{ssec:monde_arch}) with the \amove strategy
(Section~\ref{ssec:monde_offload}). We also exploit the opportunity for expert
load balancing by assigning the hot/cold experts to more workload-friendly
hardware (Section~\ref{ssec:monde_balancing}).
\section{\name} \label{sec:monde}
This section presents Mixture of Near-Data Experts (\name{}), a CXL-enabled memory
system with custom near-data computing capabilities for MoE FFN inference.
The key idea of \name{} is to store and process large expert parameters within
the CXL-based memory. Prior works move large parameters to the GPU for computation.
In contrast, \name{} takes a different approach by \emph{transferring activations
to where parameters reside} and performing expert computation on-site, which
significantly reduces the burden of data movement.

\subsection{\name Device Architecture} \label{ssec:monde_arch}
Figure~\ref{fig:monde_overview}(a) shows an overview of the \name{} system. This
subsection introduces each internal component of the \name{} device.

\keyword{CXL \& NDP Controllers.}
The CXL controller (\ding{182}) manages the CXL protocols for communicating with
the host system. The CXL controller identifies host-invoked NDP instructions
wrapped in the CXL \emph{Request with Data} (RwD) messages by using the NDP flag
defined in the reserved bits of a message flit. The NDP instructions are forwarded
and queued to the internal memory-mapped instruction buffer of the \name{} NDP
controller. The NDP controller (\ding{183}) generates memory requests and NDP
requests that load data tiles into the NDP core, and triggers expert computations.
Once the computation is finished, the NDP controller stores the output activations
in the designated device memory address and signals the host by setting the
memory-mapped \emph{done} register. 

\keyword{MoNDE NDP Core.}
The \name{} NDP core (\ding{184}) processes the MoE experts located in the
device memory.
Our design exploits the expert skew to efficiently process \emph{fat and wide} matrix
multiplications of cold experts.
An expert computation is essentially a matrix multiplication of
$\mathrm{Output\ } C=\mathrm{Input\ }A \times \mathrm{Expert\ }B$, where matrices
A and C correspond to the input and output activations, while B corresponds
to the expert parameter.
For cold experts, matrices A and C have small height dimensions due to the small
number of routed tokens. In contrast, the width dimensions, either $d_{model}$ or
$d_{ff}$, are constantly large, often multiples of 256.
In order to maintain high compute utilization for small token counts, our
NDP core design adopts small-height 4$\times$4 multiply-and-accumulate (MAC)
processing element (PE) arrays. We use 64 of such arrays that are controlled
by a SIMD controller.
Using the aforementioned architecture, the \name NDP core processes 4$\times$256
matrix operations in a consecutive tile-by-tile, output-stationary manner.

\keyword{Device Memory.}
We refer to a corporate CXL memory device~\cite{samsung_cxl_pnm} to build on a
realistic DRAM module (\ding{185}).
\name uses LPDDR SDRAM, which uses wide-I/O technology for high memory
bandwidth and low power.
Each $\times$16 chip has a total of 16 Gb density and a maximum
transfer rate of 8533 MT/s. Each DRAM module is composed of 32 chips that
provide in total, 64 GB of memory capacity and 68 GB/s of bandwidth.
By utilizing 8 memory channels, the \name memory device allows access to 512 GB
of memory capacity and approximately 512 GB/s of bandwidth.

\subsection{Offload and Fetch Strategy} \label{ssec:monde_offload}
The MoE Transformers consist of the unconditionally-used \emph{dense}
parameters and the MoE expert parameters that are dynamically and only
conditionally activated by the gating function. Because the latter have massive
sizes that often do not fit into a single GPU system, we choose to offload all
expert parameters to the \name{} memory device, while keeping the dense
parameters in the GPU memory as in~\cite{se-moe}.
The existing LLM offloading frameworks~\cite{deepspeed} need to transfer the
corresponding expert parameters from the CPU memory for MoE computation and
evict other experts from the GPU.
We refer to this as the \emph{\pmoveFull} (\pmove) strategy.
In contrast, we propose the \emph{\amoveFull} (\amove) strategy in which MoE
expert operations are processed using the \name NDP core by transferring the
input activation for experts from the GPU to the \name{} device and returning
the output of expert computations back to the GPU afterwards.
Figure~\ref{fig:monde_overview}(b) shows the overview of the proposed \amove as
compared to \pmove.

\keyword{Analytical Comparison of \pmove and \amove.}
We formulate Equations~\ref{eq:pswap_complexity} and~\ref{eq:aswap_complexity}
to show the data movement complexities of the two strategies.
Typically, MoE Transformers scale towards $d_{model}$, $d_{ff}$ and $E$,
thereby making \pmove scale in a \emph{cubic} fashion~\cite{switchtransformers}.
Moving such massive amount of data from the memory device to the GPU on-demand
will incur long latency overheads since the PCIe bandwidth is limited.
In contrast, the data volume scaling of \amove can be reduced to $O(d_{model})$
when $B$ and $S$ are small, which is the case for many MoE inference
tasks. As illustrated in Figure~\ref{fig:moti_analyses}(b), there is a significant
gap between the data volumes of \pmove and \amove for LLMs that scale towards the
expert size and embedding dimensions.
{
\small
\begin{align} 
  \mathrm{Parameter\ Movement} & = 2\times E\times d_{model}\times d_{ff} 
  \label{eq:pswap_complexity}\\ 
  \mathrm{Activation\ Movement} & = 2\times B\times S\times d_{model} 
  \label{eq:aswap_complexity}
\end{align}
}

\subsection{GPU-\name{} Load Balancing} \label{ssec:monde_balancing}
We propose GPU-\name{} load-balancing, which leverages the two hardware to
concurrently process different expert operations and reduce the execution time
for MoE layers. We exploit the imbalance in expert load and the computing power
of the hardware units to assign workloads to each hardware based on the compute
and memory intensity of each expert. Our algorithm assigns the top-$H$
compute-intensive hot experts to the GPU and the remaining experts to the \name{}
NDP, and overlaps the GPU (\pmove-to-expert) and \name NDP (\amove-to-expert)
execution.

The $H$ value sensitively affects performance, as assigning too many experts to
the GPU can cause excessive data movement, whereas the opposite can underutilize
GPU resources. Our goal is to find the $H$ value that balances the runtime of
the GPU and \name workflows ($t_{GWF}$ and $t_{MDWF}$) shown in
Equation~\ref{eq:runtimes}. We use two intuitions for determining $H$. First,
the GPU computation latency $t_{GPU}$ and \amove latency $t_{AM}$ are
negligibly small for inference and thus are removed from consideration.
Second, we approximate the \pmove latency $t_{PM}$ and \name NDP computation latency
$t_{MD}$, as shown in Equation~\ref{eq:runtimes2}, considering their
bandwidth-bound nature. Here, $Expert_{GPU}$ and $Expert_{MD}$ each represent
the expert parameters processed on each hardware, and $BW_{PCIe}$ and $BW_{MD}$
represent the PCIe and the \name device memory bandwidth.

Equation~\ref{eq:expert_cnt} denotes the number of activated experts
$Expert_{Activ}$, that is, experts with at least 1 input token.
Lastly, equating the formulas in Equation~\ref{eq:runtimes2} and applying
Equation~\ref{eq:expert_cnt}, we find the $H$ value as
Equation~\ref{eq:h_formula}.
$H$ is computed during runtime after the gating function. We use the bandwidth
value from the hardware specification, but this can be replaced by profiled
bandwidths. We add a scaling factor $\alpha$ to micro-control $H$ for when
the overall NDP experts have increased compute intensity, making our second
intuition invalid. In such cases, increasing $H$ to offload more experts to
the GPU workflow reduces end-to-end latency. Finding the scaling factor is
untrivial, as many factors (e.g., tokens-per-expert, embedding size, GPU
compute power) need to be considered collectively. Inspired by auto-tuning
features used by~\cite{deepspeed}, the \name framework auto-tunes the scaling
factor by periodically running profiled inference on a small set of past input
batches and finding the local optima among $H$ candidates (e.g., $H+1$, $H+2$).
{
    \small
    \begin{align}
    & t_{GWF} = t_{PM} + t_{GPU} & t_{MDWF} = t_{AM} + t_{MD}&\label{eq:runtimes} \\
    & t_{PM} \approx \frac{Expert_{GPU}}{BW_{PCIe}} & t_{MD} \approx \frac{Expert_{MD}}{BW_{MD}}&\label{eq:runtimes2}
    \end{align}%
    \vspace{-0.14in}
    \begin{gather}
    Expert_{Activ} = Expert_{GPU} + Expert_{MD} \label{eq:expert_cnt} \\
    H = \alpha \times Expert_{GPU}  = \alpha{}\times{}\frac{BW_{PCIe}}{BW_{MD}+BW_{PCIe}}\times{}Expert_{Activ} \label{eq:h_formula}
    \end{gather}%
}

For multi-\name device scenarios, the \name algorithm obtains the $H$ value by using
the \emph{aggregate} \name device bandwidth. The NDP units are load-balanced
by distributing expert workloads sorted by compute intensity in a round-robin manner.
The expert input activations are separately transferred to each \name device, after
which each \name NDP device processes the given input data. The output activations
are retrieved from each \name device to the GPU sequentially for the MoE combine
operation.

\vspace{-0.07in}
\subsection{Programming Model}

\vspace{-0.03in}
\keyword{Host Interface.}
\name adopts a heterogeneous programming model (e.g., CUDA) in which the host
launches a kernel and the NDP device executes the offloaded instruction. The
host does this through the host-side \name device driver, which generates and
offloads \name NDP instructions via the CXL interface. The \name NDP controller
raises the memory-mapped \emph{done register} once a kernel execution is completed.
We define two kernels: \texttt{gemm} and \texttt{gemm+relu}. The \texttt{gemm}
kernel offloads an expert GEMM instruction to the \name NDP. The \texttt{gemm+relu}
kernel is an extension that runs a tailing activation function (i.e., ReLU or GeLU).
A host kernel is compiled into a 64-Byte CXL instruction, which includes a
4-bit opcode (including reserved ops), a 48-Byte (address, data size)
metadata of the input/output activation and expert parameters, and auxiliary
NDP flags such as \texttt{isNDP} for identifying NDP instructions.

\keyword{Memory Allocation.}
The host-side device driver allocates fixed-sized memory space for the MoE expert
parameters and input/output activations in the \name device memory during MoE layer
initialization. Data in the \name memory space is mapped to the DRAM \emph{ro-ba-bg-ra-co-ch},
in order to fully utilize the DRAM bandwidth for contiguous memory accesses.
To mitigate memory contention from accessing expert parameters and activations
simultaneously, we map each data in different \emph{banks}: the parameters and
activations are each mapped to the even and odd-indexed banks.

\keyword{Execution Flow.}
We demonstrate the execution flow of a \name expert operation. First, the input
activation data is transferred to the \name memory with \amove. Once the input
activations had been written to device memory, the host-side device driver issues
and queues gemm instructions in the memory-mapped instruction buffer at the \name
NDP controller, which are decoded to generate device-side memory and NDP requests.
The \name memory quickly populates the \name NDP scratchpad and operand buffers
with expert parameter and input activation \emph{tiles}. The tiled operands are
reshaped into skewed formats and processed by the systolic array, after which the
output activation tiles are written to the designated output memory space.
Finally, the NDP controller raises the memory-mapped \emph{done} flag.
\vspace{-0.05in}
\section {Evaluation} \label{sec:eval}

\vspace{-0.03in}
\subsection{Experimental Setup} \label{ssec:setup}

\setlength{\aboverulesep}{1.3pt}
\setlength{\belowrulesep}{2.6pt}
\begin{table}[t]
    \vspace{-0.10in}
    \caption{Workloads and system configurations.}
    \vspace{-0.10in}
    \label{tab:workloads}
    
   \resizebox{0.98\columnwidth}{!}{
        \begin{tabular}{ccccc}
        
            \toprule[1.2pt]
            \multirow{2}{*}{\textbf{Model}}  & \textbf{Non-Expert}   & \textbf{Expert}            & \multirow{2}{*}{\textbf{$d_{model}$}}  & \multirow{2}{*}{\textbf{$E$}}  \\ 
                                             & \textbf{Params (GB)}  & \textbf{Params (GB)}       &                               &                       \\
            \midrule
            Switch-Large-128                 & 1.1                   & 51.5              & 1024                          & 128           \\
            \midrule
            NLLB-MoE                         & 5.7                   & 103.1             & 2048                          & 128           \\
            
            \toprule[1.2pt]
            \textbf{Model}      & \textbf{Gating}           & \multicolumn{3}{c}{\textbf{Task}}                             \\
            \midrule
            Switch-Large-128    & top-1                     & \multicolumn{3}{c}{XSum Language Modeling}                    \\
            \midrule
            NLLB-MoE            & top-2                     & \multicolumn{3}{c}{FLORES-200 Machine Translation}            \\

            \bottomrule[1.2pt]
            \\
            
            \toprule[1.2pt]
            \textbf{Platform}       & \multicolumn{4}{c}{\textbf{System Configuration}} \\
            \midrule
            CPU                     & \multicolumn{4}{c}{Intel Xeon Silver 4310 CPU, 187 GB/s Memory Bandwidth} \\
            \midrule
            GPU                     & \multicolumn{4}{c}{$1\times$ NVIDIA A100 GPU PCIe} \\
            \midrule
            \multirow{4}{*}{MoNDE}  & \multirow{2}{*}{Compute}  & \multicolumn{3}{c}{64 units of 4$\times$4 Systolic Array} \\ 
                                    &                           & \multicolumn{3}{c}{264 KB Buffers @ 1 GHz} \\
                                    \cmidrule{2-5}
                                    & Memory   & \multicolumn{3}{c}{512 GB/s Bandwidth, 512 GB Capacity} \\
            \midrule
            Interconnect            & \multicolumn{4}{c}{PCIe Gen4 $\times$16} \\ 
            \bottomrule[1.2pt]
            
            \end{tabular}
    }
    \vspace{-0.20in}
\end{table}

\noindent \textbf{\name NDP Model.}
To evaluate the \name workflow, we first use the NVIDIA Nsight profiler to
obtain a detailed MoE latency breakdown and isolate the latency for expert
computation for the evaluated models.
We then implement a cycle-level expert computation simulator for which we use
Ramulator~\cite{ramulator} to model our \name memory.
Based on the number of tokens routed to each \name-offloaded expert, which is
determined at runtime, the simulator outputs the latency required for the \name
NDP to process the same expert computations.
Finally, we replace only the CPU computation latency obtained from the profiler
with the NDP computation latency obtained from the cycle-level simulator to
estimate the \name inference latency.
Overall, the GPU performs Transformer operations while fetching the expert
parameters from the CPU memory or offloading expert computations to the CPU.
We focus on the single-GPU setting enhanced by CXL-expanded memory with and
without NDP support, because we aim to replace costly GPU resources with more
affordable NDP-enabled memory devices.

\keyword{Implementation.}
We implement the \name operation flows using Pytorch APIs.
We modify the Hugging Face MoE model implementations~\cite{hf-transformers}
to implement a drop-less and padding-less token routing algorithm similar
to~\cite{meta-moe-inf}.
We implement on-demand \pmove~\cite{meta-moe-inf}, in which only the
\emph{activated experts} are fetched to the GPU, instead of over-fetching
the entire experts as in~\cite{deepspeed,se-moe}.
We use the PyTorch CUDA memory copy API for \amove between the GPU and CPU for
expert computation on the CPU, which we use to model the \name NDP behavior.
Lastly, we implement the GPU-\name load-balancing algorithm in our codebase.

\vspace{0.05in}
\noindent \textbf{Workloads.}
Table~\ref{tab:workloads} summarizes our workloads and system configurations.
We evaluate \name{} using the pre-trained Switch Transformers and NLLB-MoE
provided in the Hugging Face repository~\cite{hf-switch-repo,nllb-moe-repo}.
For both models, we run the 128-expert MoE models, which are typically hard to
fit in a single commercial GPU.
We use the bfloat16 datatype, which is widely-adopted for inference tasks.
We evaluate the encoder and decoder performance individually, as \name can be
applied to any encoder-only~\cite{bert} or decoder-only~\cite{gpt4} MoE LLMs.
We use the input sequence length of 512 for each batch.

\subsection{Performance} \label{ssec:eval-speedup}
\begin{figure}[t]
    \centering
    \vspace{-0.10in}
    \includegraphics[trim=4.8in 3.3in 4.8in 3.3in, clip=True,
        width=\columnwidth]{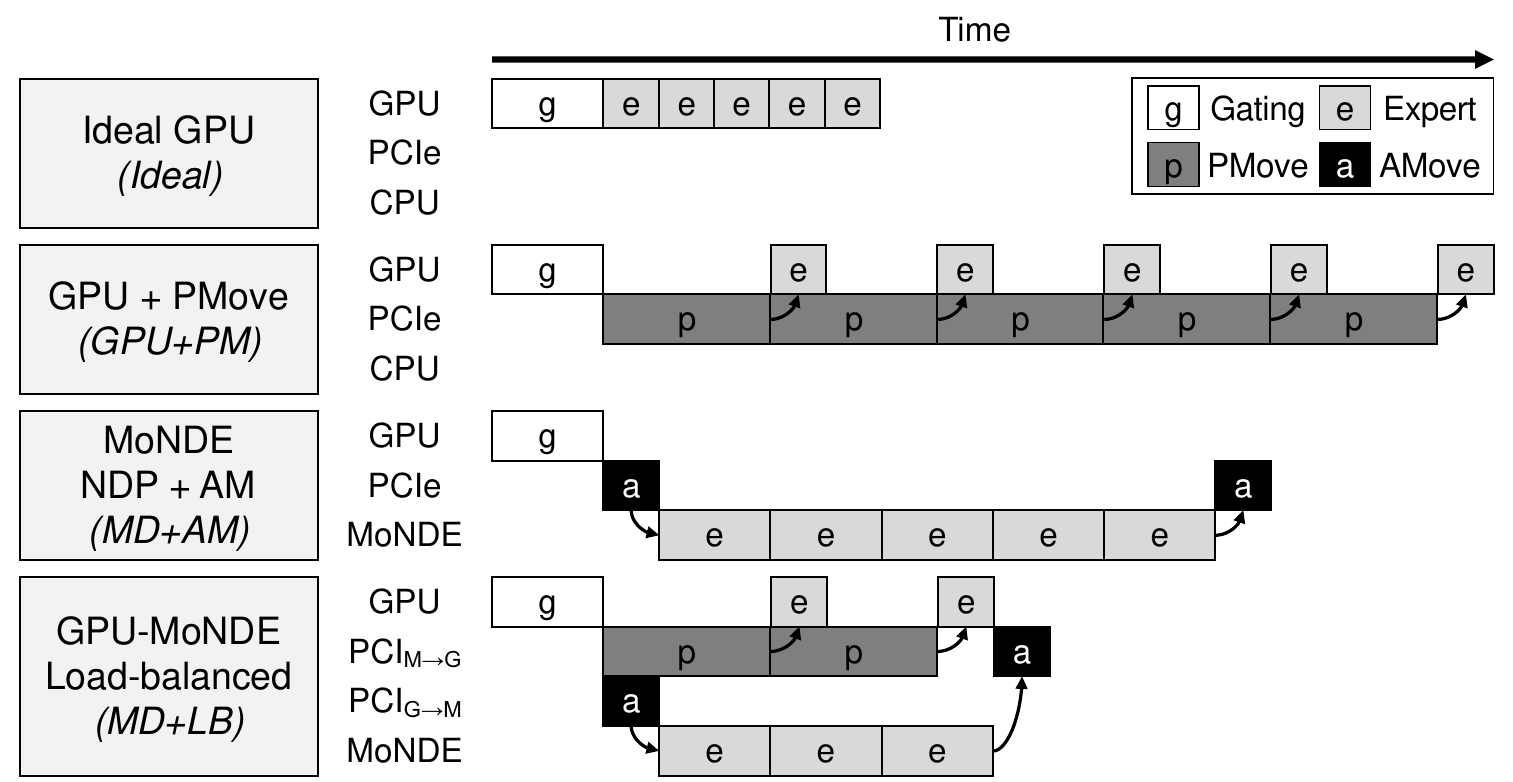}
    \vspace{-0.20in}
    \caption{
      Comparsion between MoE workflows. 
    }
    \vspace{-0.10in}
    \label{fig:opt_overview}
\end{figure}

We compare the following configurations.
The GPU with \pmove{} support (GPU+PM) is the memory offloading scheme where
activated MoE experts are moved to the GPU for computation.
The \name NDP (MD+AM) scheme runs all expert operations using the \name{} NDP.
The input activations are \amove{}d between the GPU and \name device.
The GPU-\name load-balanced scheme (MD+LB) uses the \name load-balancing
algorithm to collaboratively use the GPU and \name NDP. Both \pmove and \amove
are used. Lastly, the ideal single-GPU (Ideal) models a GPU with infinite memory
capacity, where all MoE and non-MoE layers reside in the GPU memory.
Figure~\ref{fig:opt_overview} depicts the workflow of MoE Transformer block
execution schemes with regard to parallel hardware streams.

\begin{figure}[t] 
    \centering
    \includegraphics[trim=4.9in 4.60in 4.85in 4.70in, clip=True,
                        width=\columnwidth]{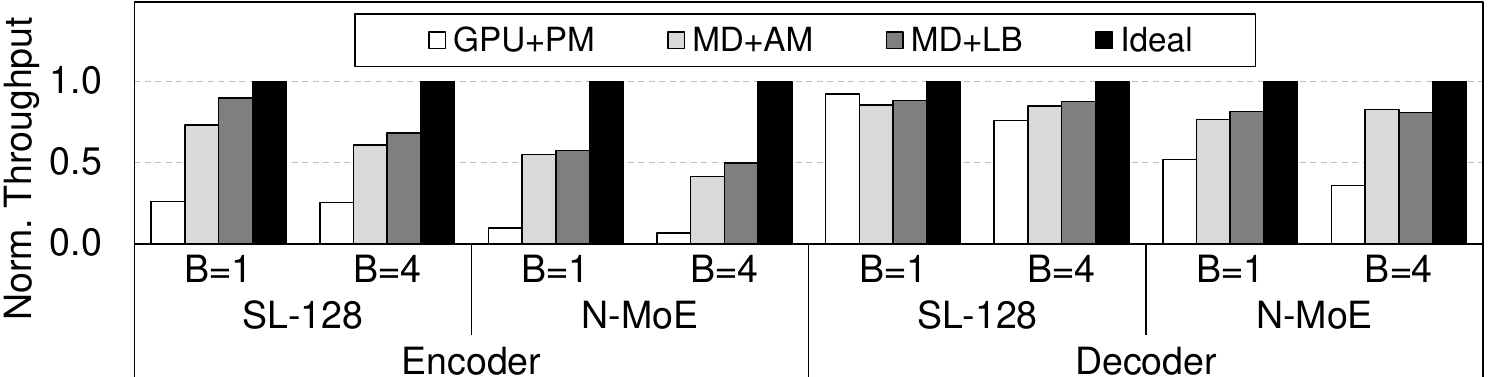}%
    \vspace{-0.10in}
    \caption{
        Normalized end-to-end throughput.
    }
    \vspace{-0.10in}
    \label{figs_eval:eval_throughput}
\end{figure}

\begin{figure}[t] 
    \centering
    \includegraphics[trim=4.9in 4.60in 4.85in 4.60in, clip=True,
                        width=\columnwidth]{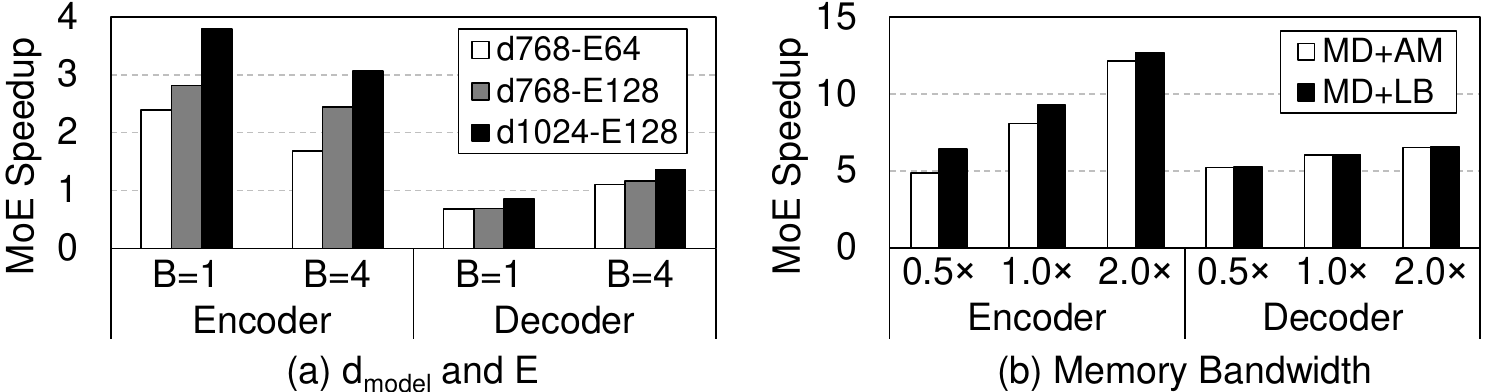}%
    \vspace{-0.10in}
    \caption{
        Sensitivity study.
    }
    \label{figs_eval:sensitivity}
    \vspace{-0.15in}
\end{figure}

\begin{figure*}[t]
\vspace{-0.10in}
\minipage{0.32\textwidth}
    \includegraphics[trim=6.5in 4.6in 6.5in 4.5in, clip=True,
                        width=\columnwidth]{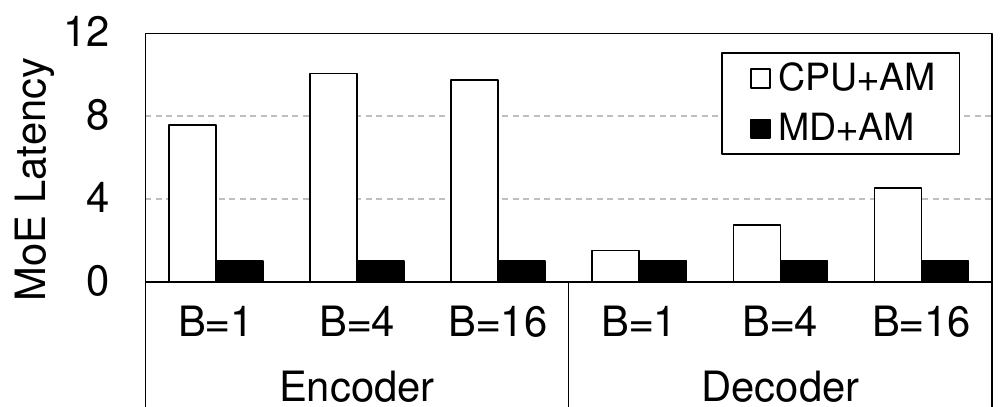}
    \vspace{-0.23in}
    \caption{Comparison with the CPU.}
    \label{figs_eval:cpu}
\endminipage\hfill
\minipage{0.32\textwidth}
    \includegraphics[trim=6.5in 4.6in 6.5in 4.5in, clip=True,
                        width=\columnwidth]{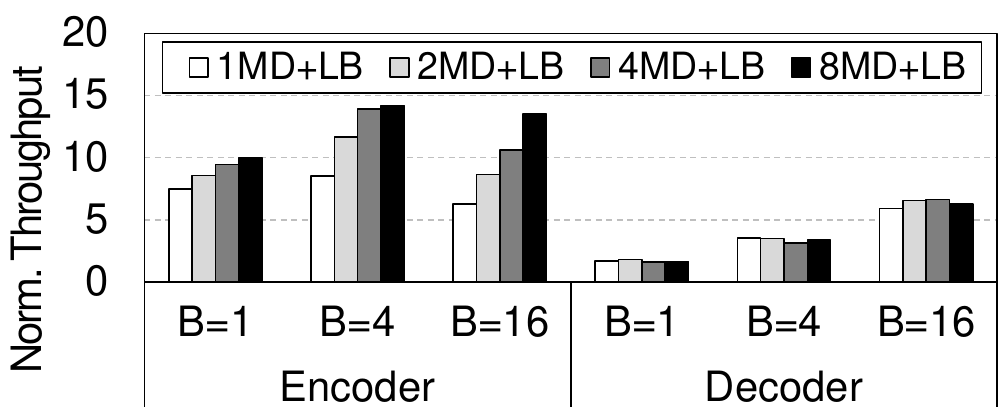}
    \vspace{-0.23in}
    \caption{
        Multi-\name evaluation.
    }
    \label{figs_eval:multi_monde}
\endminipage\hfill
\minipage{0.32\textwidth}%
    \includegraphics[trim=6.5in 4.6in 6.5in 4.5in, clip=True,
                    width=\columnwidth]{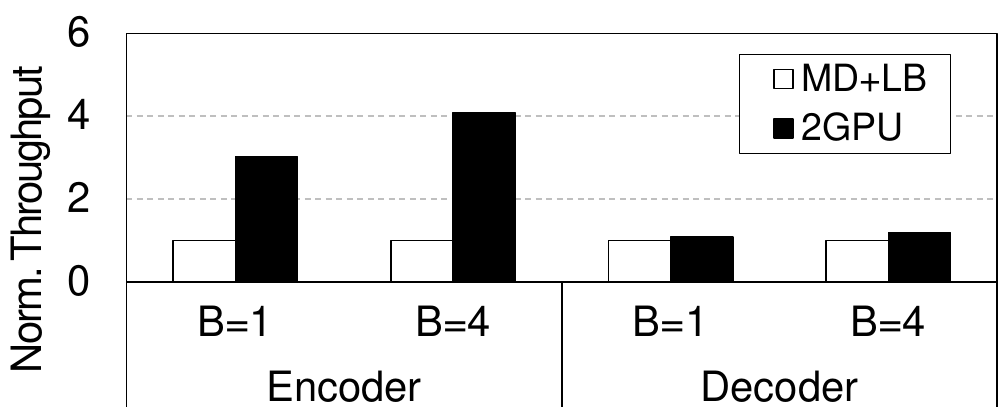}
    \vspace{-0.23in}
    \caption{
        Comparison with multi-GPU.
        \label{figs_eval:multi_gpu}
    }
\endminipage
\vspace{-0.10in}
\end{figure*}

\keyword{End-to-End Throughput.}
Figure~\ref{figs_eval:eval_throughput} shows the \name throughput normalized to
the Ideal scenario across different batch sizes. The results show that the MD+LB
scheme improves encoder and decoder throughput over GPU+PM by 3.1$\times$
and 1.1$\times$ for Switch-Large (SL-128), and by 6.7$\times$ and 1.9$\times$ for
NLLB-MoE (N-MoE) on average.
The performance benefit of MD+AM comes from avoiding the long \pmove latency by
replacing it with small \amove. MD+LB improves upon MD+AM by leveraging both
the GPU and \name NDP, achieving an average speedup (across SL-128 and N-MoE)
of $4.9\times$ and $1.5\times$ over GPU+PM for the encoder and decoder.

\keyword{Sensitivity.}
Figure~\ref{figs_eval:sensitivity}(a) shows the speedup of MD+LB over GPU+PM
for three variants of Switch Transformers with different $d_{model}$ and $E$
configurations.
MD+LB shows increasingly higher speedups for larger models, reflecting its
robustness to $d_{model}$ and $E$ scaling.
Figure~\ref{figs_eval:sensitivity}(b) shows the speedups of MD+AM and MD+LB
over GPU+PM for NLLB-MoE (batch-4) with 0.5$\times$-2.0$\times$ the \name memory
bandwidth and rate-matching NDP compute.
For both the encoder and decoder, the speedups increase since higher memory
bandwidth leads to latency reduction for cold experts, which comprise
the majority of the MoE experts.
The MD+LB constantly shows better performance over MD+AM by adaptively
controlling the $H$ value to utilize both the GPU and \name NDP for expert
computation. Higher memory bandwidth leads to lower and more conservative $H$
value, which explains why the gap between the two policies is reduced.
We see smaller gains for the decoder because only a small number of experts are
activated.

\keyword{Comparison with the CPU.}
Figure~\ref{figs_eval:cpu} compares the MoE latency of CPU expert
computation (CPU+AM) and MD+LB for NLLB-MoE.
MD+AM shows an average of $9.1\times$ and $1.9\times$ latency reductions for the 
encoder and decoder, which can be attributed to higher \name memory
bandwidth ($2.7\times$) than the CPU memory.
Even with higher CPU memory bandwidth, however, fully utilizing the CPU memory
is often challenging due to remote NUMA accesses, which can degrade CPU
performance.
Furthermore, the CPU performance is not scalable, whereas \name performance can
be scaled by adding more \name devices via the PCIe slots for improved throughput.

\keyword{Scalability.} 
Figure~\ref{figs_eval:multi_monde} presents the multi-\name inference
throughput for the MoE layers of NLLB-MoE, which we normalize to GPU+PM.
For the encoder, employing more \name devices improves performance, largely
due to the increase in compute power and memory bandwidth.
For the decoder, the performance gain over GPU+PM is similar across the
numbers of \name devices within each batch size because the small number
of input tokens (i.e., 1/4/16) cannot fully utilize multiple \name NDP units.

\keyword{Comparison with Multi-GPU.} 
With MoE expert parallelism, expert parameters can be distributed
across multiple GPUs to fit in the GPU memory~\cite{deepspeed}. However,
multi-GPU systems are inefficient when serving MoE decoders as they are
auto-regressive; each input token activates only one or two experts,
and the GPUs with inactive experts remain idle.
Figure~\ref{figs_eval:multi_gpu} compares the throughput between MD+LB
and a 2-GPU setting for the encoder and decoder of NLLB-MoE.
The multi-GPU system shows a higher throughput for the encoder due to a larger
number of activated experts in each GPU, whereas for the decoder, \name shows a
throughput comparable to the multi-GPU system.
Because a single \name device provides memory capacity that is comparable to
dozens of modern GPUs, \name is more cost-effective in serving generative LLMs.

\begin{table}[b]
    \vspace{-0.10in}
    \caption{Summary of \name{} area and power.}
    \vspace{-0.10in}
    \label{tab:eval-hw}
    \centering
    \resizebox{0.95\columnwidth}{!}{%
        \begin{tabular}{|c|ccc|c|}
        \hline
        \multirow{2}{*}{Component}   & \multicolumn{3}{c|}{Systolic Array}                                         & \multirow{2}{*}{Scratchpad} \\ \cline{2-4}
                                     & \multicolumn{1}{c|}{PE}    & \multicolumn{1}{c|}{Control} & Operand Bufs    &                             \\ \hline
        Area ($mm^2$)                & \multicolumn{1}{c|}{2.042} & \multicolumn{1}{c|}{0.053}   & 0.289           & 0.570                       \\ \hline
        Power (W)                    & \multicolumn{1}{c|}{0.993} & \multicolumn{1}{c|}{0.033}   & 0.258           & 0.526                       \\ \hline
        \end{tabular}
   }
\end{table}

\subsection{Area and Power Consumption} \label{ssec:eval-hw}
Table~\ref{tab:eval-hw} presents the area and power of the \name{} NDP core. We use
Synopsys Design Compiler to synthesize the \name{} systolic array with a 28 nm
technology node at 1 GHz clock. We also generate on-chip buffers with a
commercial memory compiler using the same technology.
Our \name NDP design adds 3.0 $mm^2$ of area overhead, which corresponds to
approximately 0.9 Gb DRAM cells of our target memory.
We estimate the power consumption of the base memory expander device to which
we apply the \name NDP unit, by using Micron DDR4-3200 power calculator and
scaling to our target LPDDR device with operating voltage. The estimation shows
that our memory device consumes 114.2 W and our NDP unit incurs only 1.6\% of
power overhead to the base memory system. 
\vspace{-0.05in}
\section{Conclusion} \label{sec:conclusion}
This paper explores Mixture of Near-Data Experts (\name) for enhancing
MoE LLM inference through near-data processing. By replacing massive
MoE expert movement invoked by data offloading techniques with
cheap activation movement and processing MoE experts near-the-data,
\name significantly reduces MoE inference latency.
When collaborating with the GPU to concurrently process MoE expert
computations that are suited for each compute hardware, \name{} achieves
inference latency comparable to an ideal GPU system with infinite
memory. Evaluation on MoE LLMs shows up to a 7.5$\times$ end-to-end
inference speedup over a strong baseline that implements the latest
parameter offloading technique.

\vspace{-0.05in}
\begin{acks}
This work was supported in part by Institute of Information \& Communications
Technology Planning \& Evaluation (IITP) grants funded by the Korean government
(MSIT) (No. 2021-0-00863 and RS-2023-00256081) and a research grant from SK Hynix.
The Institute of Engineering Research at Seoul National University provided
research facilities for this work, and the EDA tool was supported by the IC
Design Education Center (IDEC), Korea.
Jaewoong Sim is the corresponding author.
\end{acks}

\vspace{-0.05in}
\bibliographystyle{ACM-Reference-Format}
\bibliography{refs}

\end{document}